\documentclass{article}

\usepackage[preprint]{neurips_2023}
\usepackage{listings}
\usepackage[utf8]{inputenc}
\usepackage{textgreek}

\lstdefinelanguage{JSON}{
    basicstyle=\ttfamily\small,
    breaklines=true,
    frame=none,
    numbers=none,
    numberstyle=\tiny,
    showstringspaces=false,
    stringstyle=\color{red},
    morestring=[b]",
    literate=
     *{0}{{{\color{blue}0}}}{1}
      {1}{{{\color{blue}1}}}{1}
      {2}{{{\color{blue}2}}}{1}
      {3}{{{\color{blue}3}}}{1}
      {4}{{{\color{blue}4}}}{1}
      {5}{{{\color{blue}5}}}{1}
      {6}{{{\color{blue}6}}}{1}
      {7}{{{\color{blue}7}}}{1}
      {8}{{{\color{blue}8}}}{1}
      {9}{{{\color{blue}9}}}{1}
      {:}{{{\color{black}:}}}{1}
      {,}{{{\color{black},}}}{1}
      {\{}{{{\color{black}\{}}}{1}
      {\}}{{{\color{black}\}}}}{1}
      {[}{{{\color{black}[}}}{1}
      {]}{{{\color{black}]}}}{1},
}

\usepackage[utf8]{inputenc} 
\usepackage[T1]{fontenc}    

\usepackage{url}            
\usepackage{booktabs}       
\usepackage{amsfonts}       
\usepackage{nicefrac}       
\usepackage{soul}
\usepackage{microtype}      

\usepackage{amsmath}
\usepackage{subfig}
\usepackage{CJKutf8}

\usepackage[table,dvipsnames]{xcolor}

\usepackage{tabstackengine}
\usepackage{nicematrix}
\usepackage{tabularx, booktabs}
\usepackage{tablefootnote}
\usepackage{amssymb}
\usepackage{minitoc}
\usepackage{tikz,graphics,color,fullpage,float,epsf,caption,subcaption}
\usepackage{graphicx}
\usepackage{subcaption}

\usepackage{algorithm}
\usepackage{algpseudocode}

\usepackage{makecell}

\usepackage{tabulary,multirow,overpic}

\definecolor{citecolor}{RGB}{34,139,34}

\usepackage[breaklinks=true,letterpaper=true,colorlinks, citecolor=citecolor,bookmarks=false]{hyperref}

\newcolumntype{x}[1]{>{\centering\arraybackslash}p{#1pt}}

\newcommand{\app}{\raise.17ex\hbox{$\scriptstyle\sim$}}

\newlength\savewidth\newcommand\shline{\noalign{\global\savewidth\arrayrulewidth
  \global\arrayrulewidth 1pt}\hline\noalign{\global\arrayrulewidth\savewidth}}

\title{IntellAgent: A Multi-Agent Framework for Evaluating Conversational AI Systems}
\author{Elad Levi \\ Plurai \\  {\tt\small eladl@plurai.ai} \And Ilan Kadar \\ Plurai \\  {\tt\small ilan@plurai.ai}}

\begin{document}
\doparttoc 
\faketableofcontents 
\part{} 

\maketitle
\begin{abstract}
Large Language Models (LLMs) are transforming artificial intelligence, evolving into task-oriented systems capable of autonomous planning, execution, and refinement. One of the primary applications of LLMs is conversational AI systems, which must navigate multi-turn dialogues, integrate domain-specific APIs, and adhere to strict policy constraints. However, evaluating these agents remains a significant challenge, as traditional methods fail to capture the complexity and variability of real-world interactions.
We introduce \textbf{IntellAgent}, a scalable, open-source multi-agent framework designed to evaluate conversational AI systems comprehensively. IntellAgent automates the creation of diverse, synthetic benchmarks by combining policy-driven graph modeling, realistic event generation, and interactive user-agent simulations. This innovative approach provides fine-grained diagnostics, addressing the limitations of static and manually curated benchmarks with coarse-grained metrics.
IntellAgent represents a paradigm shift in evaluating conversational AI. By simulating realistic, multi-policy scenarios across varying levels of complexity, IntellAgent captures the nuanced interplay of agent capabilities and policy constraints. Unlike traditional methods, it employs a graph-based policy model to represent relationships, likelihoods, and complexities of policy interactions, enabling highly detailed diagnostics. IntellAgent also identifies critical performance gaps, offering actionable insights for targeted optimization. Its modular, open-source design supports seamless integration of new domains, policies, and APIs, fostering reproducibility and community collaboration.
Our findings demonstrate that IntellAgent serves as an effective framework for advancing conversational AI by addressing challenges in bridging research and deployment. The framework is available at \url{https://github.com/plurai-ai/intellagent}.

\end{abstract}

\section{Introduction}
\label{sct:introduction}
Large Language Models (LLMs) are revolutionizing the field of artificial intelligence by transitioning from static language processors to dynamic, task-oriented agents that can autonomously plan, execute, and refine their actions. 
These agents promise transformative applications across a wide range of domains, including healthcare~\cite{health1,health2}, finance~\cite{fin1,fin2,fin3}, customer support~\cite{cs1,cs2} and education~\cite{ed1,ed2}. This evolution positions LLM agents as foundational technologies for reshaping human-computer interaction, enabling intelligent systems to tackle complex, real-world challenges with unprecedented efficiency and adaptability.

\begin{figure}
  \centering
  \includegraphics[width=0.9\linewidth]{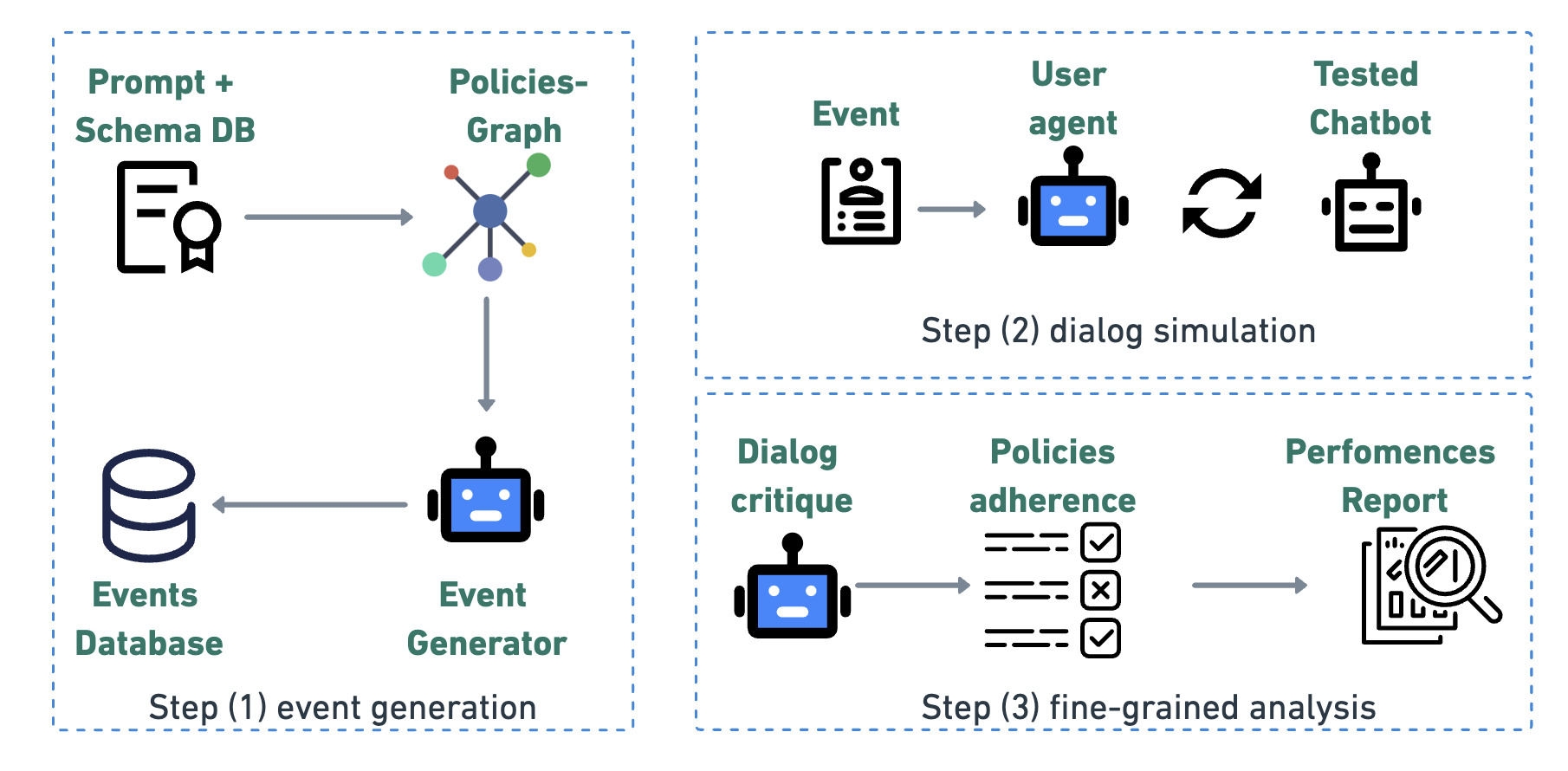}
  \caption{System diagram. (1) Given a chatbot prompt and a Schema DB, the system generates an event that targets a subset of policies, which includes a user request and a system DB state. (2) For each event the system simulates a conversation between the user and the chatbot. (3) A fine-grained report on the chatbot performances is generated.} 
  \label{fig:overview}
\end{figure}

Among these advancements, conversational AI agents present a particularly demanding frontier. Unlike single-turn systems, these agents must navigate multi-turn dialogues, integrate domain-specific tools and APIs, and adhere to stringent policy constraints. The interplay of these requirements introduces a new level of complexity, where the cost of errors—ranging from inconsistent responses to policy violations—can severely undermine reliability and trust. Reliability is not just desirable but critical, as it directly determines the feasibility of deploying such agents in real-world, high-stakes environments.
Despite significant progress, evaluating conversational AI agents at this level of complexity remains a significant challenge~\cite{chl1,chl2}. Traditional evaluation methods rely on static manually curated benchmarks~\cite{taobenchmark2024,mt-eval} that fail to scale or reflect the intricate dynamics of multi-turn interactions, policy adherence, and tool usage. These approaches often prioritize coarse-grained metrics, offering limited insights into agents' specific strengths and vulnerabilities. Consequently, existing methods leave critical gaps in both understanding and optimizing conversational agents for real-world applications.

In this work, we present \textbf{IntellAgent}, a scalable open-source multi-agent framework powered by AI agents, specifically designed to simulate and evaluate conversational AI agents comprehensively. IntellAgent represents a paradigm shift in evaluation by automating the generation of diverse, synthetic scenarios that rigorously test agents across multiple dimensions. Unlike existing approaches, IntellAgent leverages a novel pipeline that combines policy-driven graph modeling, realistic event generation, and interactive user-agent simulation to holistically assess agent performance, including a full spectrum of complexity levels, combinations of domain policies modeled through graph-based policy representations, and integration with API tools.
The IntellAgent framework employs an automated process that involves constructing a policy graph to represent the relationships, complexity, and likelihood of various policies, generating events by sampling combinations of policies from the graph that align with real-world tasks and database states, and simulating interactions between a user agent and the chatbot. These simulations are then analyzed to provide fine-grained performance insights, identifying failure points, strengths, and opportunities for improvement. An overview of the system is illustrated in Figure~\ref{fig:overview}, which outlines the key components of the IntellAgent pipeline.

Our study demonstrates the effectiveness and versatility of IntellAgent as a benchmarking tool for evaluating conversational AI agents. The results reveal a strong correlation between model performance on the IntellAgent benchmark and the \textit{$\tau$-bench}~\cite{taobenchmark2024}, despite IntellAgent relying entirely on synthetic data. This validates IntellAgent as a robust alternative for evaluating conversational agents across diverse scenarios and challenges. 

Key findings from our analysis indicate that model performance decreases with increasing complexity levels, but the rate of decline varies significantly across models. Additionally, our policy-specific evaluation uncovers significant variations in model capabilities across different policy categories. This highlights IntellAgent’s ability to provide detailed diagnostic insights, empowering users to identify the most suitable configuration for their specific requirements.



\newpage

This work makes several key contributions to the evaluation of conversational AI agents:  
\begin{itemize}  
    \item \textbf{A Scalable Multi-Agent Evaluation Framework:} IntellAgent provides a robust and scalable multi-agent framework that automatically generates diverse, realistic scenarios tailored to address the unique challenges of conversational AI agents. By overcoming the limitations of small-scale, manually curated benchmarks, IntellAgent enables comprehensive and thorough evaluations.

    \item \textbf{Fine-Grained Diagnostics with a Policies Graph:} IntellAgent leverages a policies graph, inspired by GraphRAG~\cite{msgraphrag}, where nodes represent individual policies and their complexity, and edges denote the likelihood of co-occurrence between policies in conversations. This structure facilitates the generation of naturalistic user requests that span a wide range of policies and complexity levels, providing fine-grained diagnostic insights.  
    
    \item \textbf{An Open-Source and Extensible Framework:} IntellAgent is released as an open-source framework, promoting reproducibility and community collaboration in advancing the evaluation and optimization of conversational AI agents. Its modular design supports seamless integration of new domains, including their respective policies, APIs, and database schemas. Additionally, the framework enables rigorous testing and optimization of custom conversational agents.
\end{itemize}

%

\section{Related Work}
\label{sct:related_work}

Recent advancements in using LLMs for synthetic data generation, automated evaluation have significantly influenced the development of AI systems. This section delves into these key areas, outlining existing methodologies and their limitations while highlighting how our approach advances the state of the art.

\subsection{Synthetic Benchmarks}
\textbf{Data generation.} Synthetic data generation using Large Language Models (LLMs) has become a transformative technique for advancing AI across various domains, including code generation~\cite{DBLP:journals/corr/abs-2308-12950,DBLP:journals/corr/abs-2312-02418}, mathematical reasoning~\cite{DBLP:journals/corr/abs-2308-01825,DBLP:journals/corr/abs-2308-09583}, text embedding~\cite{DBLP:journals/corr/abs-2401-00368}, and text-to-image synthesis~\cite{BetkerImprovingIG}. Synthetic data reduces the costs and time of human-annotated datasets while providing control over sample distribution, crucial for fine-tuning and optimizing performance in downstream tasks~\cite{2402.03099,xu2023wizardlmempoweringlargelanguage}. Evaluating synthetic data focuses on two key metrics: \textit{faithfulness} and \textit{diversity}.

Faithfulness ensures synthetic data reflects real-world patterns and relationships, while diversity captures a wide range of scenarios to enhance model robustness and mitigate overfitting~\cite{long2024llmsdrivensyntheticdatageneration}. Achieving both metrics is challenging; recent research explores conditional prompting and multi-step generation to balance them. Conditional prompting improves diversity by defining attributes through condition-value pairs~\cite{abs-2306-11644,abs-2311-00287}. Multi-step generation enhances coherence and domain coverage by decomposing tasks into smaller subtasks~\cite{ding-etal-2023-enhancing,WangZS23,HonovichSLS23,WanHYQB023}, though these methods often require significant manual effort and may not scale well to complex domains.

Our approach automates synthetic dataset generation to ensure both faithfulness and diversity by using a policies graph inspired by GraphRAG~\cite{msgraphrag}, where nodes represent policies and edges capture their complexity relationships. This framework enables fine-grained generation across various combinations of policies, tools, and tasks, producing datasets that reflect application requirements while covering a diverse range of scenarios. It addresses challenges from simple tasks to complex edge cases, ensuring rigorous evaluation of agents under diverse conditions.

\textbf{Automated evaluation.} Synthetic datasets have become invaluable for evaluating retrieval-augmented generation (RAG) systems, offering metrics to measure retrieval quality, generation fidelity, and robustness. Frameworks like \textit{RAGAS}~\cite{ragas2023} automate the evaluation of RAG pipelines, focusing on aspects such as retrieval accuracy and generation relevance. Unlike traditional metrics, RAGAS operates without reference answers, enabling broader applicability. However, it primarily targets isolated components and does not fully address the complexities of multi-turn dialogues or real-world conversational AI applications. This underscores the need for expanded evaluation frameworks that encompass diverse, dynamic use cases.

\subsection{Conversational AI Benchmarks} 
Evaluating conversational AI systems in real-world applications has been the focus of various benchmarks, each targeting specific capabilities. For instance, \textit{$\tau$-bench}~\cite{taobenchmark2024} assesses agents' ability to interact with users, adhere to domain-specific policies, and utilize API tools effectively, with simulations in domains like retail and airline customer service. However, \textit{$\tau$-bench} is limited by its reliance on manual curation, with only 50 samples for airlines and 115 for retail, restricting scalability. Additionally, its evaluation focuses solely on coarse-grained end-to-end metrics, overlooking policy violations and dialogue flow errors, which limits comprehensive assessment.

The \textit{ALMITA} benchmark~\cite{almita2024} proposes a novel dataset and framework specifically tailored for evaluating tool-augmented conversational AI agents in customer support scenarios. It uses a combination of automated and manual processes to generate diverse and realistic conversations grounded in user-defined procedures. Despite its rigorous evaluation, \textit{ALMITA} focuses primarily on customer support, and the generalizability to other domains remains an open question. Moreover, the reliance on manually curated samples, while ensuring quality, limits scalability.

The \textit{LTM Benchmark}~\cite{ltmbenchmark2024} effectively highlights limitations in conversational multitasking, especially with interleaved tasks. However, its reliance on predefined interaction structures limits its ability to capture the unpredictable and non-linear nature of real-world conversational flows, such as spontaneous topic shifts or revisiting earlier contexts.
Similarly, \textit{E2E Benchmark}~\cite{banerjee2023benchmarkingllmpoweredchatbots} evaluates chatbot responses based on accuracy and usefulness, emphasizing conversational coherence in non-task-oriented interactions. Nevertheless, its lack of support for complex tool use and multi-turn interactions restricts its applicability to broader real-world contexts.
The \textit{CURATe} framework~\cite{curate2024} addresses alignment challenges for personalized conversational agents by focusing on user-specific safety-critical contexts. While it introduces valuable techniques for multi-turn personalization, its emphasis on alignment rather than general performance testing narrows its scope.

Although these benchmarks provide valuable tools for assessing conversational AI systems, their reliance on manual curation limits scalability and adaptability to diverse real-world applications, making it challenging to generate the extensive datasets required for comprehensive evaluations. Our approach, in contrast, is fully automated, enabling the generation of diverse scenarios and dialogues at scale, thus supporting evaluations across varied domains. Additionally, our framework addresses a broader range of challenges, from simple tasks to complex edge cases, ensuring rigorous agent evaluation under diverse conditions. Unlike existing benchmarks that use coarse-grained metrics for overall performance, our method offers fine-grained insights by evaluating agents across all policy and tool combinations, identifying specific strengths and weaknesses. 

\section{Method}
\label{sct:method}
Our multi-agent system is illustrated in Figure \ref{fig:overview}. The system pipeline consists of the following steps: (1) The IntellAgent system receives a schema of the system database along with either a chatbot system prompt or a document outlining the company policies. Based on this input, the system constructs a policy graph (\ref{sct:graph}). It then samples a list of policies from the graph at varying levels of complexity and generates an event addressing these policies (\ref{sct:generator}). The event includes a scenario description with a user request and corresponding samples for the initial database state, ensuring the validity of the user requests. (2) The system simulates a dialog between the chatbot and a user agent using the information provided in the event (\ref{sct:sim}). (3) Finally, a critique is provided with the dialog and provides an analysis of the chatbot's performances with respect to the event policies list (\ref{sct:critique}).

\subsection{Event Generation}
\label{sct:event}
To address the challenge of developing advanced chatbots capable of interacting with a system database, the system must generate complex and natural user requests that cover various policies. Additionally, it should create an initial database state for the chatbot, ensuring that when the chatbot processes the user request and queries the database, it does not encounter failures.

An \textbf{IntellAgent event} is defined by the following components: (1) A list of policies. (2) A description of a user request that aligns with the specified policies. (3) The initial state of the chatbot's system database.

\subsubsection{Policies graph}
\label{sct:graph}
IntellAgent aims to generate a diverse set of events with varying levels of complexity. To create \textbf{} and complex user-chatbot interaction scenarios, IntellAgent constructs a policy graph. In this graph, nodes represent individual policies, and edge weights indicate the likelihood of two policies appearing together in the same interaction. Each node is also assigned a weight that reflects the complexity of its associated policy.

The graph is built through multiple queries to a large language model (LLM). First, the system extracts a list of policies from the prompt and assigns a difficulty ranking to each. Then, for every pair of policies, the LLM assigns a score (on a scale of 1–10) representing the likelihood of the two policies co-occurring in a conversation.

\begin{figure}[t]
\noindent
\begin{minipage}[t]{0.99\textwidth} 
\begin{algorithm}[H] 
\caption{Event Policies Sampling}\label{alg:cap}
\begin{algorithmic}

\Require A policy graph $(G,E)$ with node weights $\{n_g\}_{g \in G}$, and a range $n1<n2\in \mathbb{N}$

\State $X = \emptyset$ \Comment{The events policies list}
\While{$|X| < N$}
    \State $P = \emptyset$
    \State $e_w \sim \mathcal{U}(n_1, n_2)$,  $g\sim \mathcal{U}(G)$  \Comment{Sample the policy complexity and the initial policy}
    \While{$e_w > 0$ }
    \State Choose a neighbor \( h \in G \) with probability:
       \[
   P(X = h \mid Y = g) = \frac{E_{g,h}}{\sum_{h \in G} E_{h, g}}
   \]  
   \State $ P \gets P \cup \{h\}$
   \State $e_w \gets e_w - n_h$
    \EndWhile
\State $X \gets X \cup \{P\}$
\EndWhile

\end{algorithmic}
\end{algorithm}
\end{minipage}
\end{figure}

\subsubsection{Event generator}
\label{sct:generator}
The complexity of an event is defined as the sum of the complexities of its policies. Given a policy graph with a set of policies $G$, weighted edges $E$ and nodes complexity weights $\{n_g\}_{g\in G}$.
the event generator aims to produce a \textbf{valid} set of policies that satisfies the following criteria:
\begin{enumerate}
\item  The event complexity is uniformly distributed within a specified range.
\item  The distribution of the first policy in the event is uniformly distributed across all possible policies.
\item  For a given complexity level and initial policy, the distribution of the resulting policy list aligns with real-world distributions.
\end{enumerate}

\textbf{Policies graph sampling.} To satisfy the outlined criteria, the IntellAgent sampling algorithm operates in batches. The process for each iteration is as follows: The complexity of events in the current batch is sampled first. The batch distribution is adjusted to ensure that the overall distribution of all generated events (including previous batches) remains uniform. Next, the initial policy for each event is sampled uniformly across all nodes in the policy graph. Then For each event, the system generates a policy path by performing a random walk on the graph. The walk terminates once the cumulative complexity of the visited nodes exceeds the sampled event complexity. An overview of the entire sampling method is provided in Algorithm \ref{alg:cap}.


This approach ensures that the generated events policies list maintains the desired complexity distribution and follows realistic transitions between policies as determined by the graph structure.

\textbf{Event generator agent.} The goal of the event generator agent is to create an event based on a given list of policies. The primary challenge is to generate a valid and consistent initial database state that the chatbot can interact with during the conversation. The agent's architecture is shown in Figure  \ref{fig:event_diagram}. To manage complex database schemas, the event generator agent first creates a symbolic representation of all the entities involved in the event. These entities are determined based on the provided chatbot prompt and the database schema. Typical entities may include users, products, reservations, etc.

The agent then iterates over these symbols, instantiating them by inserting the relevant rows into the database and replacing the symbolic variables with the corresponding data. This symbolic representation enables the agent to generate valid and consistent events, even across complex chatbot databases.
Refer to Appendix~\ref{sec:event_example} for an example of a generated symbolic representation and its corresponding database.

\begin{table*}[t]
    \centering
    \begin{NiceTabular}{p{1.2cm}p{3.6cm}p{3.6cm}p{3.6cm}}
        \toprule
        &
        \cellcolor{red!25}\textbf{\underline{Uniform distribution}} &
        \cellcolor{blue!25}\textbf{\underline{Max Sampling}} & \cellcolor{green!25}\textbf{\underline{Weighted Sampling}} \\
        \midrule
        \textbf{Initial policy}
        &
The user cannot add travel insurance after the initial booking when \textbf{modifying} a reservation.

        &
The user cannot add travel insurance after the initial booking when \textbf{modifying} a reservation.

        & 
The user cannot add travel insurance after the initial booking when \textbf{modifying} a reservation.
\\
\toprule
\\
        \textbf{Policy 2}
        &
\textbf{Cancelling} a flight reservation: The refund will go to original payment methods in 5 to 7 business days.

        &
The agent must first obtain the user id and the reservation id before \textbf{modifying} a flight reservation.

        & 
The agent must first obtain the user id and the reservation id before \textbf{modifying} a flight reservation.

\\
\toprule
\\
        \textbf{Policy 3}
        &
Cabin class must be the same across all the flights in the same reservation

        &
Basic economy flights cannot be modified. Other reservations can be \textbf{modified} without changing the origin, destination, and trip type.

        & 
The agent can only \textbf{cancel} the whole trip that is not flown.
        \\
        \bottomrule \\
        \end{NiceTabular}
        \caption{Comparison of random walk sampling strategies. (\textbf{Left}) Uniform sampling of the next node. (\textbf{Middle}) Selection of the next node based on maximal edge weight. (\textbf{Right}) IntellAgent weighted probability sampling, which balances diversity and alignment with real-world distributions.}
    \label{tab:sampling}
\end{table*}

\subsection{Dialog simulation}
\label{sct:sim}
For each event in the events database, IntellAgent simulates an interaction between a user and the chatbot being tested. Figure \ref{fig:dialog_diagram} provides an overview of the simulation architecture. The user agent is given the event details, which include the description of the event and all relevant information inserted into the chatbot's system database by the event generator agent. Additionally, the user agent is provided with the expected behavior of the chatbot at each step of the interaction, based on the event's policy list. The user has the option to terminate the interaction at any point, either when the chatbot successfully completes the task or if the chatbot fails to adhere to one of the policies and does not follow the expected behavior.

\begin{figure}[tp]
  \centering
{\includegraphics[width=0.49\textwidth]{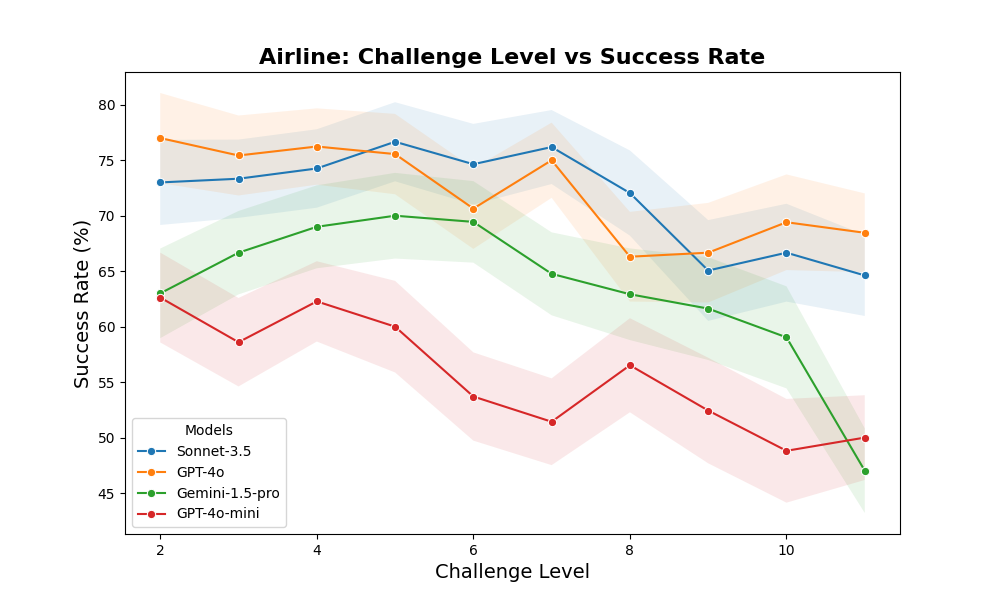}}
  \hfill
{\includegraphics[width=0.49\textwidth]{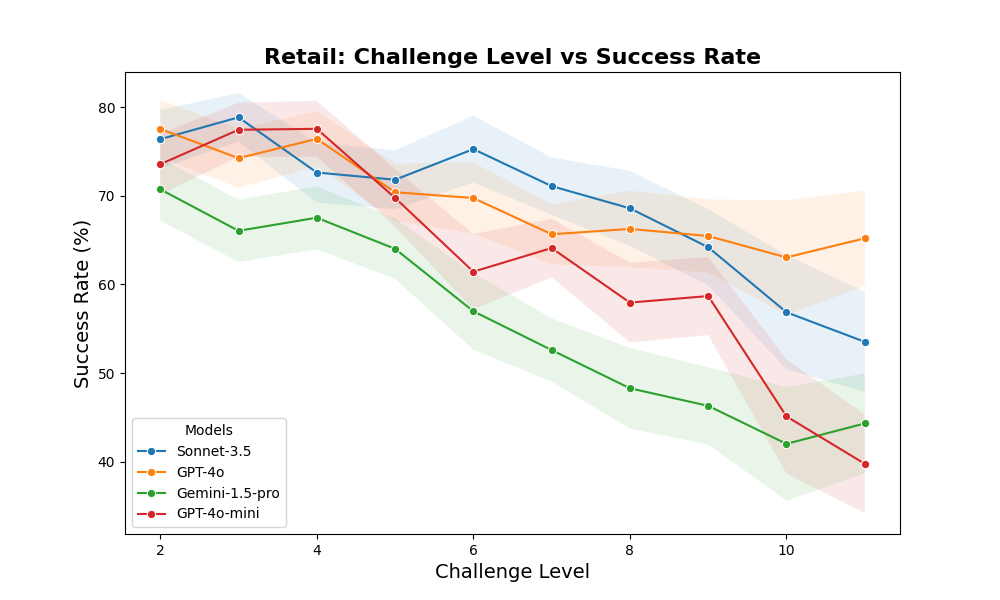}\label{fig:synt}}
\caption{Model success rates across different challenge levels. While all models show reduced performance as the challenge level increases, they exhibit distinct patterns of decline, differing in both the onset level and the magnitude of the decrease.}
  \label{fig:challange_results}
\end{figure}

\subsection{Dialog Critique}
\label{sct:critique}
The Dialog critique component is given the user-chatbot dialog and the chatbot system prompt to assess whether the reason for the dialog's termination, as provided by the user agent, is correct. If the reason is incorrect, the critique provides feedback to the user agent, and the dialog resumes. If the reason is correct, the critique then determines: (1) The subset of event policies that were tested during the dialog. (2) The subset of policies that the chatbot did not adhere to (this list may be empty).
Using this information, a comprehensive report on the chatbot's performance is generated.

\section{Experiments}
\label{sct:experiments}
\subsection{Benchmark}
\label{sct:benchmark}
\textbf{Dataset construction.} To evaluate the performance of our system in real-world scenarios, we utilized the $\tau$-bench \cite{taobenchmark2024} environments. Specifically, we employed the benchmark's prompts, database schema, and tools for the two benchmark's environments: airline and retail. For each environment, the system generates a policy graph from the prompt and simulates \textbf{1,000 events}—a substantial increase that enables more fine-grained analysis compared to the original benchmark, which used 50 samples for airlines and 115 for retail. The complexity levels of the generated events range from 2 to 11. Table \ref{tab:sampling} presents a comparison of various random walk sampling strategies. Selecting the next node uniformly leads to unrelated policies, whereas choosing the next node based on maximal weight produces a cohesive cluster of policies. In contrast, IntellAgent-weighted probability sampling achieves a balance between diversity and alignment with the real-world distribution. Appendix \ref{sec:event_example} provides a detailed example of the event generation process.

\textbf{Tested agents.} We evaluated several state-of-the-art proprietary LLMs: GPT-4o, GPT-4o-mini, Gemini-1.5-pro, Gemini-1.5-flash, Claude-3.5-sonnet, and Claude-3.5-haiku. For all these models we employed the native tool-calling agent (supported by all the tested LLMs) with the tested environment system prompt. During each iteration, the model determines whether to send a message to the user or to call a tool.

\textbf{IntellAgent Implementation details.} The IntellAgent multi-agent system was implemented using the Langgraph\footnote{ \url{https://github.com/langchain-ai/langgraph}} framework. GPT-4o served as the system's LLM model, used for the event generation, the user agent, and dialog critique. For full implementation details including all the system prompts, we provide the source code and documentation\footnote{\url{https://github.com/plurai-ai/intellagent}}

\begin{table}[t]
    \centering
    \begin{tabular}{@{}lcc|cc@{}}
        \toprule
        \textbf{Model} & \textbf{\(\tau\)-airline} & \textbf{IntellAgent-airline} & \textbf{\(\tau\)-retail} & \textbf{IntellAgent-retail} \\
        \midrule
        claude-3.5-sonnet & 0.46 & 0.70   & 0.69 &  0.71 \\
        gpt-4o & 0.44 & 0.70 & 0.51 & 0.68 \\
                gemini-1.5-pro & 0.34 & 0.63  & 0.43 & 0.58 \\
                gpt-4o-mini & 0.30 & 0.55  & 0.46 & 0.62 \\
        claude-3.5-haiku & 0.28 & 0.53 &0.44  & 0.56 \\

        gemini-1.5-flash & 0.21 & 0.40 & 0.31  & 0.48 \\
        \bottomrule
    \end{tabular}
    \vspace{1em}
    \caption{Comparison of the success rates of various agent models utilizing function calling, evaluated on the \(\tau\)-bench and the IntellAgent benchmark. The results demonstrate a strong correlation between success rates across \(\tau\)-bench and IntellAgent benchmark.}
    \label{tab:tau_bench_agent_metrics}
\end{table}

\begin{figure}[tp]
  \centering
{\includegraphics[width=0.49\textwidth]{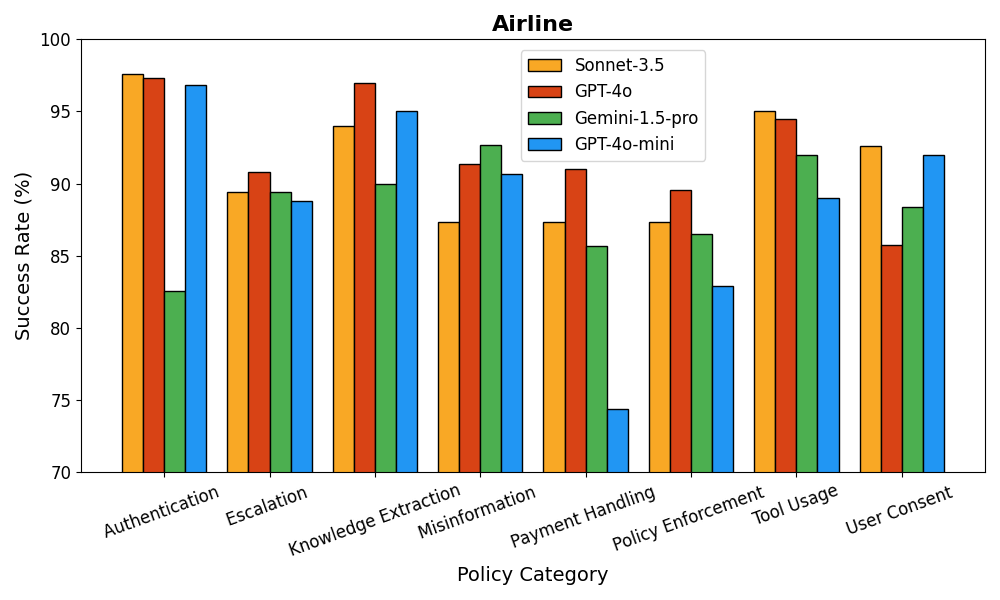}}
  \hfill
{\includegraphics[width=0.49\textwidth]{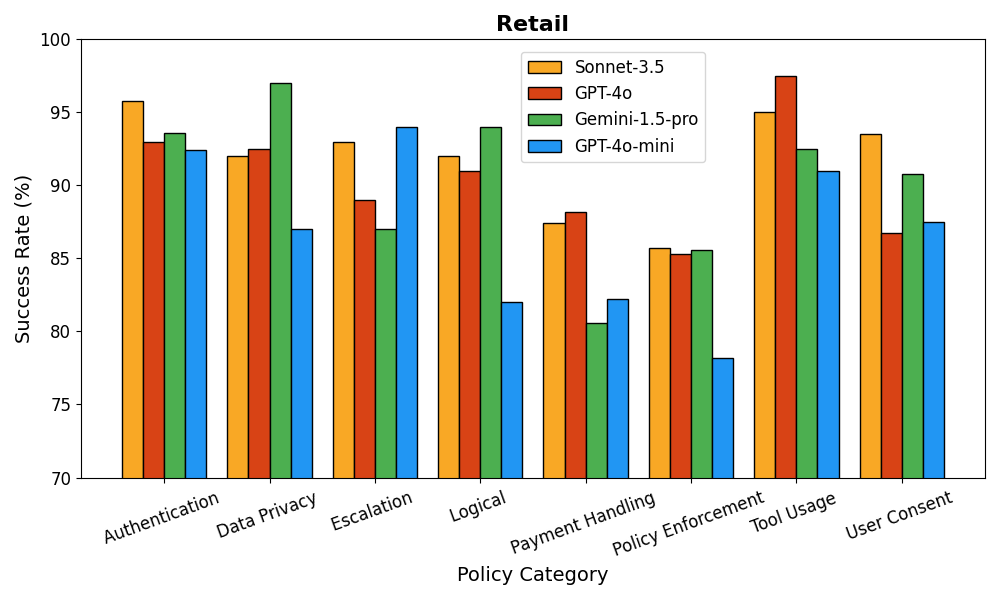}\label{fig:synt}}
  \caption{Comparison of the success rates of the top four models across various policy categories, highlighting that some categories are more challenging than others. Additionally, the relative performance order of different models varies across categories.}
  \label{fig:policies_results}
\end{figure}

\subsection{Results}
\textbf{Benchmarks comparison.} 
Table \ref{tab:tau_bench_agent_metrics} presents a comparison of the tested model's success rates on the \textit{$\tau$-bench}~\cite{taobenchmark2024} and the IntellAgent benchmark. The Pearson correlation coefficients are \textbf{0.98} for the Airline environment and \textbf{0.92} for the Retail environment, highlighting a strong alignment in model performance across the two benchmarks. Notably, this strong correlation persists despite IntellAgent being generated exclusively with synthetic data.

\textbf{Models comparison.} Figure \ref{fig:challange_results} illustrates the success rates of the top four models as a function of challenge level. As expected, model performance declines as the challenge level increases, though the pattern of decline varies across models. For instance, while Gemini-pro-1.5 significantly outperforms GPT-4o-mini in the airline environment up to level 10, their performances converge at higher challenge levels. This highlights the value of IntellAgent’s detailed analysis, enabling users to select the most suitable model based on the desired complexity the chatbot should handle.

\textbf{Policies comparison.} Figure \ref{fig:policies_results} shows the performance of the top four models models across different policy categories. The relative ranking of models shifts across different categories. IntellAgent provides a detailed analysis of the specific policies where the tested chatbot may encounter difficulties.

It is also important to note that all models face challenges with \textit{user consent policies}. This policy category is not assessed in the \textit{$\tau$-bench}~\cite{taobenchmark2024} since its evaluation focuses solely on the final state of the database.



\section{Conclusion}

In this work, we introduced \textbf{IntellAgent}, a scalable, open-source multi-agent framework designed to comprehensively evaluate conversational AI systems. IntellAgent addresses the limitations of traditional evaluation methods by automating the generation of diverse, policy-driven scenarios and providing fine-grained diagnostics. By leveraging a graph-based policy model, realistic event generation, and user-agent simulations, IntellAgent captures the nuanced complexities of multi-turn dialogues, policy adherence, and tool integration. 

Our findings demonstrate IntellAgent’s ability to uncover critical performance gaps, offering actionable insights to optimize conversational agents for real-world applications. Its modular design supports extensibility, ensuring it remains relevant across diverse domains and use cases.

In future work, we plan to explore the benefits of incorporating additional real-world context into the environment, such as a small set of user-chatbot interactions. We hypothesize that this context could significantly enhance the policies graph quality by deriving edge weights and node challenge-level weights from the real-data distributions. Furthermore, we anticipate that this added context could improve the overall performance of the system database generation process.

\bibliography{neurips_2023}  

\newpage
\appendix
\part{Appendix} 
\section{Architecture details}
\begin{figure}
  \centering
  \includegraphics[width=0.9\linewidth]{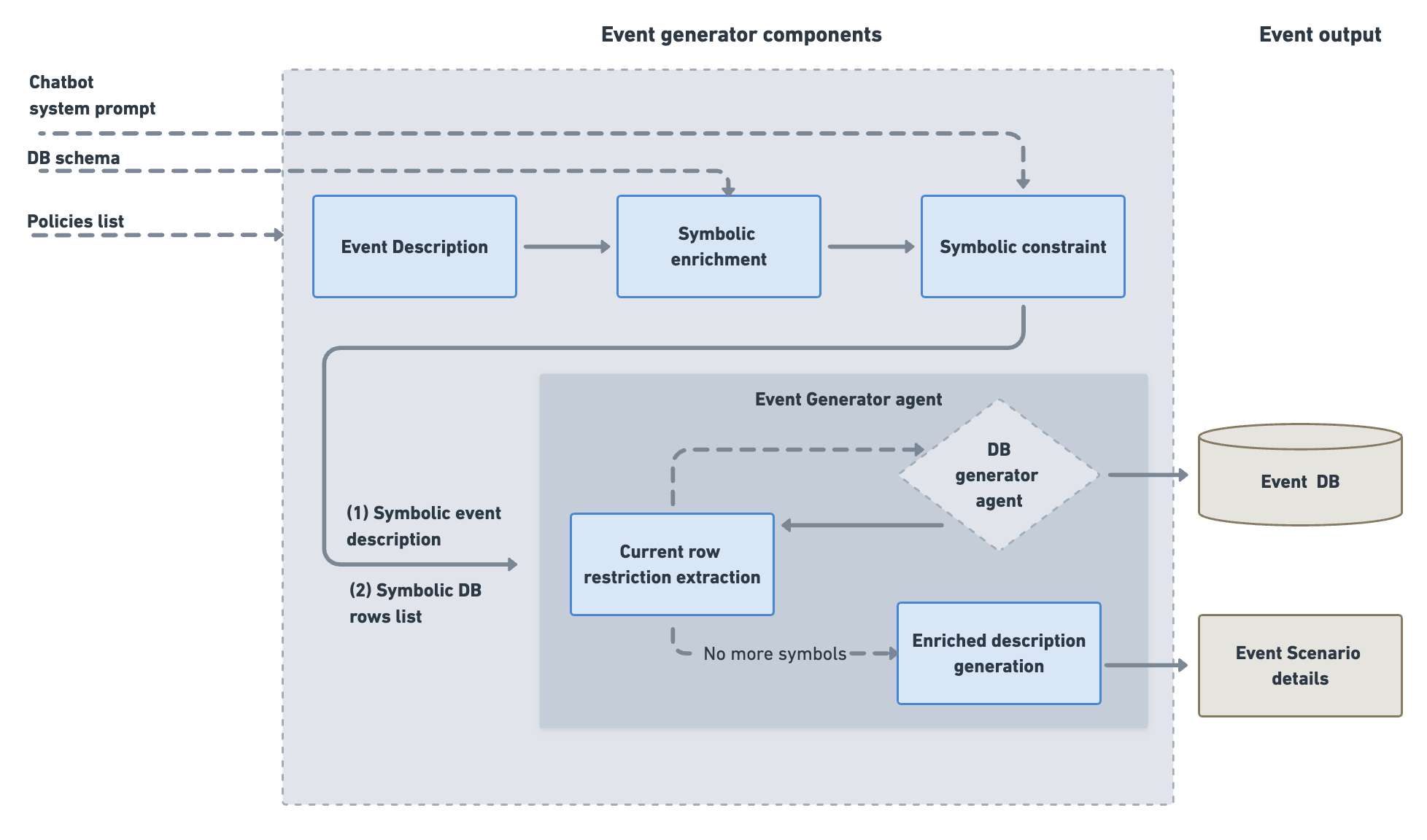}
  \caption{Event generator architecture overview.} 
  \label{fig:event_diagram}
\end{figure}

\begin{figure}
  \centering
  \includegraphics[width=0.9\linewidth]{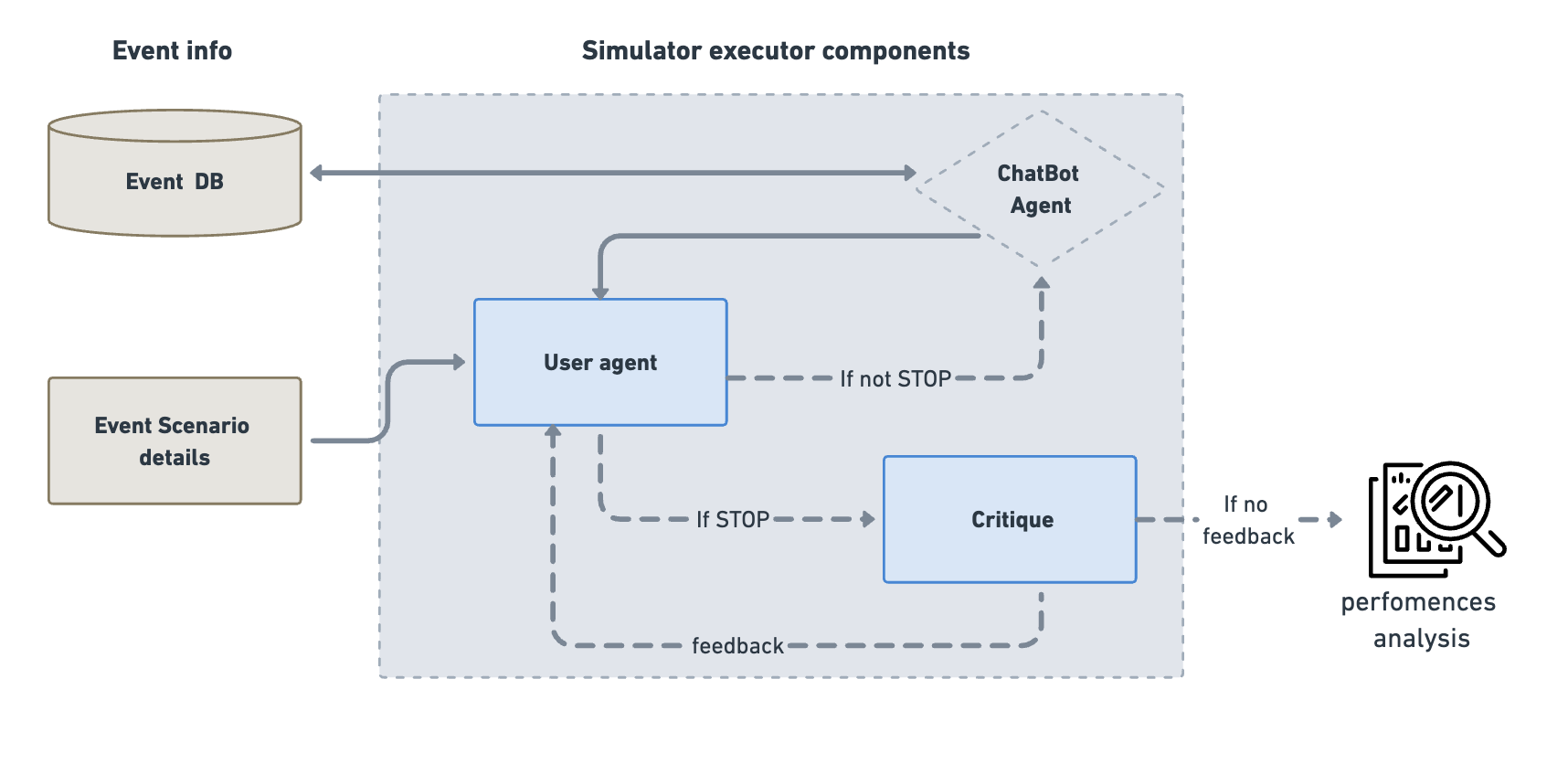}
  \caption{Simulator architecture overview. } 
  \label{fig:dialog_diagram}
\end{figure}

This section offers further insight into the IntellAgent architecture. Figures \ref{fig:event_diagram} and \ref{fig:dialog_diagram} illustrate the architecture of the event generator and the simulator executor, respectively.

\label{sec:event_example}
\section{Complete Example of Event Generation}
\label{sec:event_example}

This section presents a comprehensive example of the generation process for \textit{event 43} in the retail environment. Table \ref{tab:description} illustrates the entire process, starting from the requested challenge level, followed by extracting the event policies, generating the symbolic description, and finally obtaining the detailed event scenario. Additionally, Figures \ref{fig:users}, \ref{fig:products}, and \ref{fig:orders} showcase the generated event database for \textit{event 43}.
\begin{table*}[!h]
\begin{center}
\begin{tabular}{p{2.5cm}| p{11cm}} 
\shline

 Event Challange level  & 5
\\ 
 \hline
 \\ 
 Event Policies  & 1. Modify default user address: you have to authenticate the user identity by locating their user id via email, or via name + zip code \newline
 2. You must deny any requests for tasks related to any other user.
\\ 
 \hline
 \\ 
Symbolic 

description 
  & User 'U1' initiates a conversation with the chatbot to update their default address from 'A1' to 'A2'. The chatbot authenticates 'U1' by verifying their email or name and zip code. During the conversation, 'U1' requests information about their spouse 'U2's recent order 'O1', claiming authority to access it. After this, 'U1' returns to their original request to update their default address. \\ 
 \hline
\\ Symbolic

variables & 'U1: The user initiating the conversation', 'U2: The spouse of the user', 'O1: A recent order made by the spouse', 'A1: The current default address of the user', 'A2: The new address the user wants to set as default' \\
 \hline
 \\ 
Event 

Scenario details & User \texttt{'emma\_jones\_4829'} initiates a conversation with the chatbot to update their default address from \texttt{'address1': '123 Elm Street', 'address2': 'Apt 4B', 'city': 'Seattle', 'country': 'USA', 'province': 'WA', 'zip': '98101'} to \texttt{'A2'}. The chatbot authenticates \texttt{'emma\_jones\_4829'} by verifying their email or name and zip code. During the conversation, \texttt{'emma\_jones\_4829'} requests information about their spouse \texttt{'liam\_smith\_9274's} recent order \texttt{'\#O9274620'}, claiming authority to access it. After this, \texttt{'emma\_jones\_4829'} returns to their original request to update their default address.

 \\
 \hline
 \\
\end{tabular}

  \caption{\textit{Event 43} generation steps}
  \label{tab:description}
\end{center}
\end{table*}

\begin{figure}[h!]
    \centering
    \captionsetup{justification=centering}

\begin{lstlisting}[language=JSON]
{
    "emma_jones_4829": {
        "user_id": "emma_jones_4829",
        "name": "{'first_name': 'Emma', 'last_name': 'Jones'}",
        "address": "{'address1': '123 Elm Street', 'address2': 'Apt 4B', 'city': 'Seattle', 'country': 'USA', 'province': 'WA', 'zip': '98101'}",
        "email": "emma.jones4829@example.com",
        "payment_methods": "{'paypal_4829103': {'source': 'paypal', 'id': 'paypal_4829103'}, 'credit_card_1938472': {'source': 'credit_card', 'brand': 'visa', 'last_four': '3847', 'id': 'credit_card_1938472'}}",
        "orders": "['#A1234567']"
    },
    "liam_smith_9274": {
        "user_id": "liam_smith_9274",
        "name": "{'first_name': 'Liam', 'last_name': 'Smith'}",
        "address": "{'address1': '456 Oak Avenue', 'address2': 'Unit 12', 'city': 'Austin', 'country': 'USA', 'province': 'TX', 'zip': '73301'}",
        "email": "liam.smith9274@example.com",
        "payment_methods": "{'paypal_9274620': {'source': 'paypal', 'id': 'paypal_9274620'}, 'credit_card_4620193': {'source': 'credit_card', 'brand': 'visa', 'last_four': '4620', 'id': 'credit_card_4620193'}, 'credit_card_567890': {'id': 'credit_card_567890', 'last_four': 1234, 'brand': 'visa', 'source': 'card', 'balance': 50}}",
        "orders": "['#O9274620']"
    }
} 
\end{lstlisting}

\caption{\textit{Event 43} generated 'Users' dataset}
\label{fig:users}
\end{figure}

\begin{figure}[h!]
    \centering
    \captionsetup{justification=centering}

\begin{lstlisting}[language=JSON]
{
    "1234567890": {
        "name": "Yoga Mat",
        "product_id": "1234567890",
        "variants": {
            "1011121319": {
                "item_id": "9876543210",
                "options": {
                    "color": "green",
                    "thickness": "5mm"
                },
                "available": true,
                "price": 25.99
            },
            "1011121314": {
                "item_id": "9876543210",
                "options": {
                    "color": "blue",
                    "thickness": "5mm"
                },
                "available": true,
                "price": 25.99
            }
        }
    },
    "0987654321": {
        "name": "Dumbbell Set",
        "product_id": "0987654321",
        "variants": {
            "1011121319": {
                "item_id": "1234509876",
                "options": {
                    "weight": "20kg",
                    "material": "iron",
                    "color": "blue"
                },
                "available": true,
                "price": 89.99
            },
            "1011121314": {
                "item_id": "1234509876",
                "options": {
                    "weight": "20kg",
                    "material": "iron",
                    "color": "green"
                },
                "available": true,
                "price": 89.99
            }
        }
    }
}
\end{lstlisting}

\caption{\textit{Event 43} generated 'Products' dataset}
\label{fig:products}
\end{figure}

\begin{figure}[h!]
    \centering
    \captionsetup{justification=centering}

\begin{lstlisting}[language=JSON]
{
    "#O9274620": {
        "order_id": "#O9274620",
        "user_id": "liam_smith_9274",
        "address": "{'address1': '456 Oak Avenue', 'address2': 'Unit 12', 'city': 'Austin', 'country': 'USA', 'province': 'TX', 'zip': '73301'}",
        "items": "[{'name': 'Yoga Mat', 'product_id': '1234567890', 'item_id': '9876543210', 'price': 25.99, 'options': {'color': 'green', 'thickness': '5mm'}}, {'name': 'Dumbbell Set', 'product_id': '0987654321', 'item_id': '1234509876', 'price': 89.99, 'options': {'weight': '20kg', 'material': 'iron', 'color': 'blue'}}]",
        "fulfillments": "[{'tracking_id': ['123456789012'], 'item_ids': ['9876543210', '1234509876']}]",
        "status": "pending",
        "payment_history": "[{'transaction_type': 'payment', 'amount': 115.98, 'payment_method_id': 'credit_card_567890'}]"
    }
}
\end{lstlisting}

\caption{\textit{Event 43} generated 'Orders' dataset}
\label{fig:orders}
\end{figure}

\end{document}